\def\year{2017}\relax

\documentclass[letterpaper]{article}
\usepackage{aaai17}
\usepackage{times}
\usepackage{helvet}
\usepackage{courier}
\usepackage{graphicx}
\usepackage{amsmath}
\usepackage{algorithm}
\usepackage{algorithmic}
\usepackage{balance}

\begin{document}

\title{ VINet: Visual-Inertial Odometry as a Sequence-to-Sequence Learning Problem }
\author{ Ronald Clark\textsuperscript{1}, Sen Wang\textsuperscript{1}, Hongkai Wen\textsuperscript{2}, Andrew Markham\textsuperscript{1} \and Niki Trigoni\textsuperscript{1}\\
\textsuperscript{1}Department of Computer Science,University of Oxford, United Kingdom\\
\textsuperscript{2}Department of Computer Science,University of Warwick, United Kingdom\\
Email: \{firstname.lastname\}@cs.ox.ac.uk
}
\maketitle
\begin{abstract}
\begin{quote}
In this paper we present an on-manifold sequence-to-sequence learning approach to motion estimation using visual and inertial sensors. It is to the best of our knowledge the first end-to-end trainable method for visual-inertial odometry which performs fusion of the data at an intermediate feature-representation level. Our method has numerous advantages over traditional approaches. Specifically, it eliminates the need for tedious manual synchronization of the camera and IMU as well as eliminating the need for manual calibration between the IMU and camera. A further advantage is that our model naturally and elegantly incorporates domain specific information which significantly mitigates drift. We show that our approach is competitive with state-of-the-art traditional methods when accurate calibration data is available and can be trained to outperform them in the presence of calibration and synchronization errors.
\end{quote}
\end{abstract}

\section{Introduction}
\label{sec:intro}
A fundamental requirement for mobile robot autonomy is the ability to be able to accurately navigate where no GPS signals are available. One of the most promising approaches to achieving this goal is through the fusion of images from a monocular camera and inertial measurement unit. This setup has the advantages of being both cheap and ubiquitous and, through the complementary nature of the sensors, has the potential to provide pose estimates which are on-par in terms of accuracy with more expensive stereo and LiDAR setups. As such monocular visual-inertial odometry (VIO) approaches have received considerable attention in the robotics community \cite{gui2015review} and current state-of-the-art approaches to VIO \cite{leutenegger2015keyframe} are able to achieve impressive accuracy. However, these approaches still suffer from strict calibration and synchronization requirements.  

\begin{figure}[t!]
\centering
\includegraphics[width=1\columnwidth]{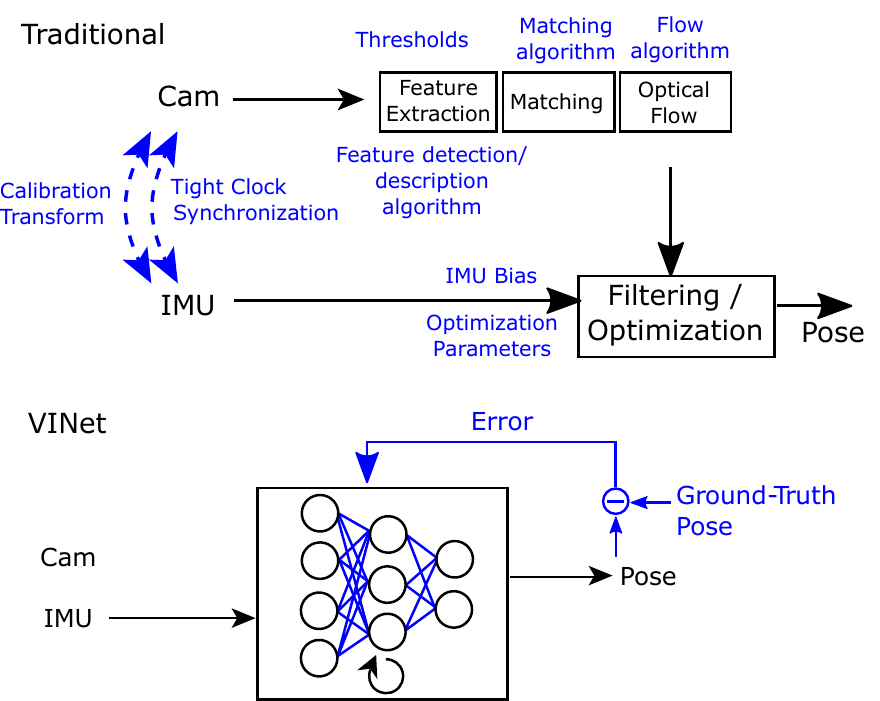}
\caption{Comparison between a standard visual-inertial odometry framework and our learning-based approach. Elements in blue need to be specified during setup. The parameters of VINet are hidden from the user and fully learned from data.}
\label{fig:hero}
\end{figure}

Inspired by the recent success of deep-learning models for processing raw, high-dimensional data, we propose in this paper a fresh approach to VIO by regarding it as a sequence-to-sequence regression problem. The resulting approach, VINet, is a fully trainable end-to-end model for performing visual-inertial odometry. Our contributions are as follows

\begin{itemize}
    \item We present the first system for visual-inertial aided navigation that is fully end-to-end trainable.
    \item We introduce a novel recurrent network architecture and training procedure to optimally train the parameters of model
    \item This includes a novel differentiable pose concatenation layer which allows the network's predictions to conform to the structure of the $SE(3)$ manifold
    \item We evaluate our method and the demonstrate its advantages over traditional methods on real-world data
\end{itemize}

\section{Related work}
In this section we briefly outline works strictly focused on monocular cameras and inertial measurement unit data in the absence of other sensors such as stereo setups and laser range finders. 

\noindent \textbf{Visual Odometry:}
VO algorithms estimate the incremental ego-motion of a camera. A traditional VO algorithm, illustrated in Fig. \ref{fig:hero} a., operates by extracting features in an image, matching the features between the current and successive images and then computing the optical flow. The motion can then be computed using the optical flow. 
The fast semi-direct monocular visual odometry (SVO) algorithm \cite{forster2014svo} is an example of a state-of-the-art VO algorithm. It is designed to be fast and robust by operating directly on the image patches, not relying on slow feature extraction. Instead, it uses probabilistic depth filters on patches of the image itself. The depth filters are then updated through whole image alignment. This visual odometry algorithm is efficient and runs in real-time on an embedded platform. Its probabilistic formulation, however, makes it difficult to tune and it also requires a bootstrapping procedure to start the process. As expected, its performance depends heavily on the hardware to prevent tracking failures - typically global shutter cameras operating at higher than 50 fps needs to be used to ensure the odometry estimates remain accurate. 

\noindent \textbf{Visual-Inertial Odometry:}
Regardless of the algorithm, traditional monocular VO solutions are unable to observe the scale of the scene and are subject to scale drift and a scale ambiguity. Dedicated loop closure methods (eg. FAB-MAP \cite{cummins2008fab}) are integrated to reduce scale drift. Scale ambiguity, however, cannot be resolved using loop-closure and requires the integration of external information. This usually takes the form of detecting the scale objects in the scene \cite{castle2010combining} \cite{pillai2015monocular} \cite{salas2013slam} or fusing information from an inertial measurement unit (IMU) to create a visual-inertial odometry (VIO) setup. Fusing inertial and visual information not only resolves the scale ambiguity but also increases the accuracy of the VO itself. In theory, the complementary nature between inertial measurements from an IMU and visual data should enable highly accurate ego-motion estimation under any circumstances: visual localization techniques are entirely reliant on observing distinctive features in the environment and in indoor situations these techniques are plagued by intermittent occlusions. On the other hand, the pose of an IMU device can be tracked through double integration of the acceleration data or more complex schemes such as the on-manifold pre-integration strategy of \cite{forster2015imu}, and then integrated with a visual-odometry method such as SVO to form a VIO system in which the IMU drift remains constrained. In state-of-the-art systems, fusion is then achieved either through a filter-based or optimization-based procedure. Filter-based methods such as the Multi-state Constraint Kalman Filter (MSCKF) \cite{mourikis2007multi}, although being more robust, consistently under-perform their optimization-based counterparts, such as the Sliding Window Filter (SWF) \cite{sibley2008sliding} and Open Keyframe VISual-inertial odometry (OK-VIS) \cite{leutenegger2015keyframe} for accuracy. In \cite{forster2015imu} it is shown that a system using the pre-integration strategy slightly outperforms OK-VIS in terms of accuracy. However, the \cite{forster2015imu} system relies on iSAM2 \cite{kaess2011isam2} as the back-end optimization and SVO as the front-end tracking system which we found fails more often than OK-VIS. We therefore compare against OK-VIS in this paper and compare against MSCKF for robustness. 

\noindent \textbf{Deep-learning:} 
Some deep-learning approaches have been proposed for visual odometry, however, to the best of our knowledge, a neural network approach has never been used in any form for end-to-end monocular visual-inertial odometry. In \cite{konda2015learning}, a Stereo-VO method is presented where they extract motion by detecting ``synchronicity'' across the stereo frames. \cite{costante2016exploring} investigated the feasibility of using a CNN to extract ego-motion from optical flow frames. In \cite{detone2016deep} the feasibility of using a CNN for extracting the homography relationship between frame pairs was shown. Recently, \cite{rambach} use an RNN to fuse the output of a standard marker-based visual pose system and IMU, thereby eliminating the need for statistical filtering.

\begin{figure*}[t!]
\centering
\includegraphics[width=\textwidth]{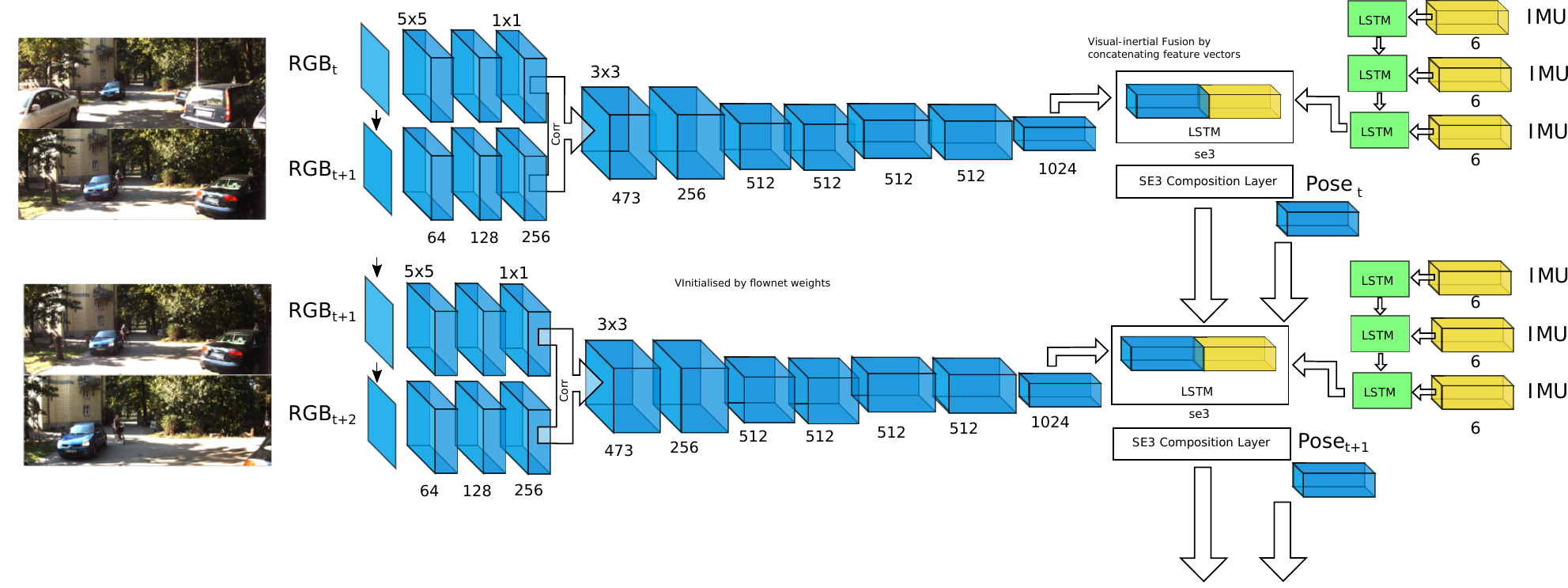}
\caption{The proposed VINet architecture for visual-inertial odometry. The network consists of a core LSTM processing the pose output at camera-rate and an IMU LSTM processing data at the IMU rate.}
\label{fig:proposed}
\end{figure*}

\section{Background: Recurrent and Convolutional Networks}
\label{sec:lstm}
\noindent Recurrent Neural Networks (RNN's) refer to a general type of neural network where the layers operate not only on the input data but also on delayed versions of the hidden layers and/or output. In this manner, the network has an internal state which it can use as a ``memory'' to keep track of past inputs and its corresponding decisions. RNN's, however, have the disadvantage that using standard training techniques they are unable to learn to store and operate on long-term trends in the input and thus do not provide much benefit over standard feed-forward networks. For this reason, the Long Short-Term Memory (LSTM) architecture was introduced to allow RNN's to learn longer-term trends \cite{hochreiter1997long}. This is accomplished through the inclusion of gating cells which allow the network to selectively store and ``forget'' memories. 
\\ \\
There are numerous variations of the LSTM architecture. However, these have been shown to have similar performance on real world data \cite{zaremba2015empirical}. The contents of the memory cell is stored in $\mathbf{c}_t$. The input gate $\mathbf{i}_t$ controls how the input enters into the contents of the memory cell for the current time-step. The forget gate, $\mathbf{f}_t$, determines when the memory cell should be emptied by producing a control signal in the range $0$ to $1$ which clears the memory cell as needed. Finally, the output gate $\mathbf{o}_t$ determines whether the contents of the memory cell should be used at the current time-step.

\begin{align} 
i_t &= \sigma(\mathbf{W}_{xi}x_t + \mathbf{W}_{hi} h_{t-1} + \mathbf{W}_{ci}c_{t-1} + b_i)    \label{eqn:forward1} \\  
f_t &= \sigma(\mathbf{W}_{xf}x_t + \mathbf{W}_{hf} h_{t-1} + \mathbf{W}_{cf} c_{t-1} + b_f)    \\  
z_t &= \mbox{tanh}(\mathbf{W}_{xc}x_t + \mathbf{W}_{hc} h_{t-1} + b_c)             \\      
c_t &= f_t \odot c_{t-1} + i_t \odot z_t                                \\         
o_t &= \sigma(\mathbf{W}_{xo} x_t + \mathbf{W}_{ho} h_{t-1} + \mathbf{W}_{co} c_t + b_o)  \\      
h_t &= o_t \odot \mbox{tanh}(c_t) \label{eqn:forward_last} 
\end{align}

Where the weights $\mathbf{W}_{:,:}$ and biases $\mathbf{b}_{:}$ fully parameterise the operation of the network and are learned during training. In order to process high-dimensional input data (such as images), convolutional layers can be integrate in the RNN structure. At each convolutional layer, multiple convolutional operations are applied to extract a number of features from the output map of the previous layer. The filter kernels with which the maps are convolved are learned during training. Specifically, the output of a convolutional layer is computed as 

\begin{equation}
y^{x,y}_{l,f} = \sum_{l,f} \mbox{act} \left( b_{l,f} + \sum_m \sum^{P_l-1}_{p=0} \sum^{Q_l-1}_{q=0} w^{p,q}_{l,f,m} y^{x+p,y+q}_{(l-1)m} \right)
\end{equation}

where $y^{x,y}_{l,f}$ is the value of the output of feature map $f$ of the $l$'th layer at location $x,y$, $b_{l,f}$ is a constant bias term also learned during training, $w^{p,q}_{l,f,m}$ is the kernel value at location $p,q$ and $P_i, Q_j$ are the width and height of the kernel.

\section{Our Approach}
Our sequence-to-sequence learning approach to visual-inertial odometry, VINet, is shown in Fig. \ref{fig:proposed}. The model consists of an CNN-RNN netowork which has been tailored to the task of visual-inertial odometry estimation. The entire network is differentiable and thus trainable end-to-end for the purpose of egomotion estimation. The input to the network is monocular RGB images and IMU data which is a 6 dimensional vector containing the $x,y,z$ components of acceleration and angular velocity measured using a gyroscope. The output of the network is a 7 dimensional vector - a 3 dimensional translation and 4 dimensional orientation quaternion - representing the change in pose of the robot from the start of the sequence. In essence, our network learns the following mapping which transforms input sequences of images and IMU data to poses 

\begin{equation}
\mbox{VIO}: \left\{\left(\mathcal{R}^{W\times H}, \mathcal{R}^6 \right)_{1:N} \right\} \rightarrow \left\{\left(\mathcal{R}^7\right)_{1:N}\right\}
\end{equation}

 Where $W\times H$ is the width and height of the input images and $1:N$ are the timesteps of the sequence. We now describe the detailed structure of our VINet model which integrates the following components.

\subsection{SE(3) Concatenation of Transformations} 
The pose of a camera relative to an initial starting point is conventionally represented as an element of the special Euclidean group $SE(3)$ of transformations. $SE(3)$ is a differentiable manifold with elements consisting of a rotation from the special orthogonal group $SO(3)$ and a translation vector, 
\begin{equation}
 \mathbf{T} = \left\{ 
 \begin{pmatrix} 
 R & T  \\0 & 1  
 \end{pmatrix} \;\bigg\vert\; R \in SO(3) \mbox{ , } T \in {\mathcal{R}}^3 
 \right\} .
\end{equation}
Producing transformation estimates belonging to $SE(3)$ is not straightforward as the $SO(3)$ component needs to be an orthogonal matrix. However, the Lie Algebra $se(3)$ of $SE(3)$, representing the instantaneous transformation,
\begin{equation}
  \frac{\xi}{dt} =\left\{ 
 \begin{pmatrix} 
 [\mathbf{\omega}]_\times & v  \\0 & 0 
 \end{pmatrix} \;\bigg\vert\; \omega \in so(3) \mbox{ , }  v \in {\mathcal{R}}^3 
 \right\},
\end{equation}
can be described by components which are not subject to orthogonality constraints. Conversion between $se(3)$ and $SE(3)$ is then easily accomplished using the exponential map

\begin{equation}
\mbox{exp:} \hspace{5pt} se(3) \rightarrow SE(3)
\end{equation}

In our network, a CNN-RNN processes the monocular sequence of images to produce an estimate of the frame-to-frame motion undergone by the camera. The CNN-RNN thus performs the mapping from the input data to the lie algebra $se(3)$. An exponential map is used to convert these to the special euclidean group $SE(3)$ where the individual motions can then be composed in $SE(3)$ to form a trajectory. In this manner, the function that the network needs to approximate remains bounded over time as the frame-to-frame motion undergone by the camera is defined by the, albeit complex, dynamics of the platform over the course of the trajectory. With the RNN we aim to learn the complex motion dynamics of the platform and account for sequential dependencies which are very difficult to model by hand.

\begin{figure}[h!]
\centering
\includegraphics[width=0.7\columnwidth]{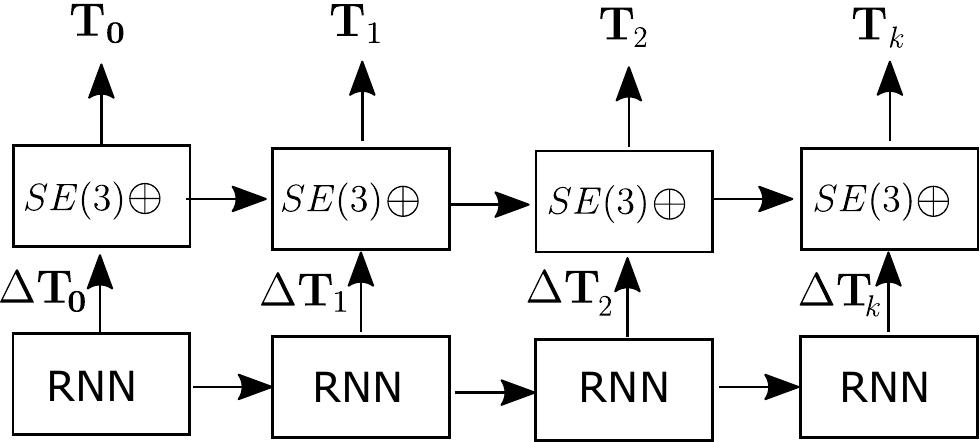}
\caption{Illustration of the SE(3) composition layer - a parameter-free layer which concatenates transformations between frames on SE(3).}
\label{fig:transformation}
\end{figure}

Furthermore, in the traditional LSTM model, the hidden state is carried over to the next time-step, but the output itself is not fed back to the input. In the case of odometry estimation the availability of the previous state is particularly important as the output is essentially an accumulation of incremental displacements at each step. Thus for our model, we directly connect the output pose produced by the $SE(3)$ concatenation layer, back as input to the Core LSTM for the next timestep. 

\subsection{Multi-rate LSTM} 
In the problem of visual-inertial odometry we are faced with the challenge of the data streams being multi-rate i.e. the IMU data often arrives at an order of magnitude (typically $10\times$) faster (100 Hz) than the visual data (10 Hz). To accommodate for this in our proposed network, we process the IMU data using a small LSTM at the IMU rate. The final hidden-layer activation of the IMU-LSTM is then carried over to the Core-LSTM.

\subsection{Optical Flow Weight Initialization} 
The CNN takes two sequential images as input and, similar to the IMU LSTM, produces a single feature-vector describing the motion that the device underwent during the passing of the two frames which is used as input to the Core LSTM. We initially experimented with two frames fed directly into a CNN pre-trained on the imagenet dataset, however, this showed incredibly slow training convergence and disappointing test performance. We therefore used as our base a network trained to predict optical flow from RGB images \cite{fischer2015flownet}. Our CNN mimics the structure of Flownet up to the Conv6 layer \cite{fischer2015flownet} where we removed the layers which produce the high-resolution optical flow output and feed in only a $1024 \times 6 \times 20$ vector which we flatten and concatenate with the feature vector produced by the IMU-LSTM before being fed to the Core LSTM. The Core-LSTM fuses the intermediate feature-level representations of the visual and inertial data to produce a pose estimate.

\subsection{Computational requirements}
The computational requirements needed for odometry prediction and the storage space required for the model are directly affected by the number of parameters used to define the model. For our network in Fig. \ref{fig:proposed}, the parameters are the weight matrices of the LSTM for both the IMU LSTM and the Core LSTM as well as the CNN network which process the images. For our network we use LSTMs with 2 layers with cells of 1000 units. Our CNN total of $55M$ trainable weights. A forward pass of images through the CNN part of the network takes on average 160ms ($\approx 10 Hz$) on a single Tesla k80. The LSTM updates are much less computationally expensive and can run at $> 200 Hz$ on the Tesla k80.

\section{Training} The entire network is trained using Backpropagation Through Time (BPTT). We use standard BPTT which works by unfolding the network for a selected number of timesteps, $T$, and then applying the standard backpropagation learning method involving two passes- a forward pass and backward pass. In the forward pass of BPTT, the activations of the network from Equations~\ref{eqn:forward1} to \ref{eqn:forward_last} are calculated successively for each timestep from time $t=1$ to $T$. Using the resulting activations, the backward pass proceeds from time $t=T$ to $t=1$ calculating the derivatives of each output unit with respect to the layer input ($x^l$) and weights of the layer ($w^l$). The final derivatives are then determined by summing over the time-steps.
Stochastic Gradient Decent (SGD) with an RMSProp adaptive learning rate is used as the to update the weights of the networks determined by the BPTT. SGD is a simple and popular method that performs very well for training a variety of machine learning models using large datasets \cite{bottou-bousquet-2008}. Using SGD, the weights of the network are updated as follows
\begin{equation}
w^l = w^l - \lambda \frac{\partial \mathcal{L}(w^l,x_t)}{\partial w^l}
\end{equation}

\noindent where $w^l$ represents a parameter (weight or bias) of the network indexed by $l$ and the learning rate ($\lambda$), which determines how strongly the derivatives influence the weight updates during each iteration of SGD. For all our training we select the best learning rate.  

\begin{figure}[h!]
\centering
\includegraphics[width=0.8\columnwidth]{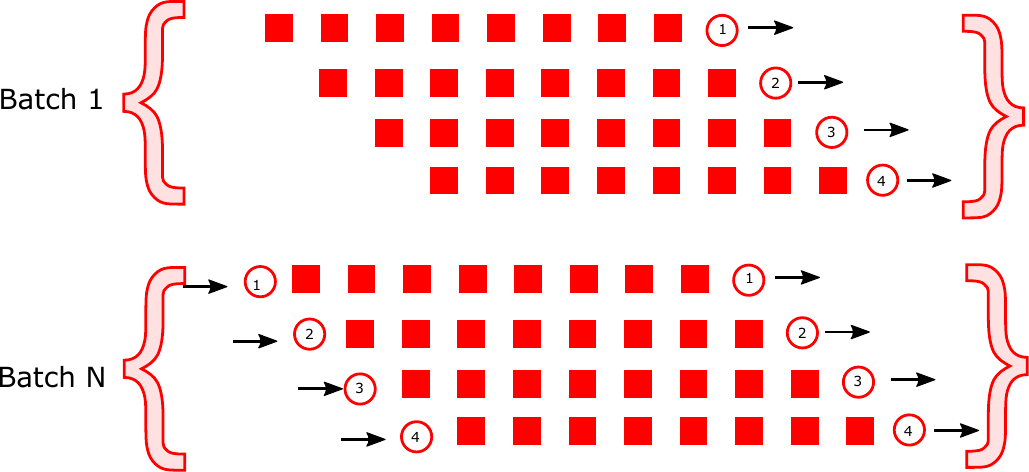}
\caption{Batch structure used for training on long odometry sequences.}
\label{fig:train_seq}
\end{figure}

Training long, continuous sequences with the high-dimensional images as input requires an excessive amount of memory. To reduce the memory required, but still keep continuity during training, we use the training structure where the training is carried out over a sliding window of batches, with the hidden state of the LSTM carried over between windows illustrated in Fig. \ref{fig:train_seq}.
Finally, we found that training the network directly through the $SE(3)$ accumulation is particularly difficult as the training procedure suffers from many local minima. In order to overcome this difficulty, we consider two losses, one based on the $se(3)$ frame-to-frame (F-2-F) predictions and the other on the $SE(3)$ full concatenated pose relative to the start of the sequence. The loss computed from the F-2-F pose is
\begin{equation} \mathcal{L}_{se(3)} = \alpha\sum ||\omega - \hat{{\omega}}|| + \beta||v - \hat{{v}}|| \end{equation} 
For full concatenated pose in $SE(3)$, we use a quaternionic representation for the orientation, giving the loss
\begin{equation} \mathcal{L}_{SE(3)} = \alpha\sum ||\mathbf{q} - \hat{{\mathbf{q}}}|| + \beta||T - \hat{{T}}||\end{equation} 
We consider three types of training; training only the $\mathcal{L}_{se(3)}$ loss, only the $\mathcal{L}_{SE(3)}$ and joint training of both losses. The weight updates for the joint training is shown in Algorithm \ref{alg:joint}.

\begin{algorithm}
\caption{Joint training of $se(3)$ and $SE(3)$ loss}
\begin{algorithmic} 
\WHILE{$i \leq n_{iter}$}
\STATE $w^{1:n} = w^{ {1:n}} - \lambda_1 \frac{\partial \mathcal{L}_{SE(3)}(w^l,x_t)}{\partial w^l}$
\STATE $w^{1:j} = w^{ 1:j}- \lambda_2 \frac{\partial \mathcal{L}_{se(3)}(w^l,x_t)}{\partial w^l}$
\ENDWHILE
\end{algorithmic}
\label{alg:joint}
\end{algorithm}
 Where $n_{iter}$ is each training iteration, $w^{{1:j}}$ are the trainable weights of the layers the SE(3) concatenation layer and $w^{{i:n}}$ are the weights of all the layers in a network with $n$ layers. During training we start with a high relative learning rate for the se(3) loss with $\lambda_2/\lambda_1 \approx 100$ and then reduce this to a very low value $\lambda_2/\lambda_1 \approx 0.1$ during the later epochs to fine-tune the concatenated pose estimation.

\section{Results}

In this section we present results evaluating the proposed method in terms of accuracy and robustness to calibration and synchronization errors and provide comparisons to traditional methods. For our experiments, we implemented our model using the Theano library \cite{bergstra2010theano} and carried out all our training on a Tesla k80. We trained the model for each dataset for 200 epochs, which took on average 6 hours per dataset. The training process did not require any user intervention, apart from setting an appropriate learning rate.

\subsection{UAV: Challenging Indoor Trajectory} 
We first evaluate our approach on the publicly-available indoor EuRoC micro-aerial-vehicle (MAV) dataset \cite{burri2016euroc}. The data for this dataset was captured using a AscTec Firefly MAV with a front-facing visual-inertial sensor unit with tight synchronization between the camera and IMU timestamps. The images were captured by a global-shutter camera at a rate of 20 Hz, and the acceleration and angular rate measurements from the IMU at 200 Hz. The 6-D ground-truth pose was captured using a Vicon motion capture system at 100 Hz. In order to provide an objective comparison to the closest related method in the literature, the optimization-based OK-VIS \cite{leutenegger2015keyframe} method is used for comparison. As we are interested in evaluating the odometric performance of the methods, no loop-closures are performed. We test the robustness of our method against camera-sensor calibration errors. We introduce calibration errors by adding a rotation of a chosen magnitude and random angle $\Delta R_{SC} \sim \mbox{vMF}(\cdot|\mu,\kappa)$ to the camera-IMU rotation matrices $R_{SC}$. For our VI Net we present two sets of results - one where we have augmented the training set by artificially mis-calibrated training data and one where only the calibrated data has been used to train the network.

\begin{figure}[h!]
\centering
\includegraphics[width=\columnwidth]{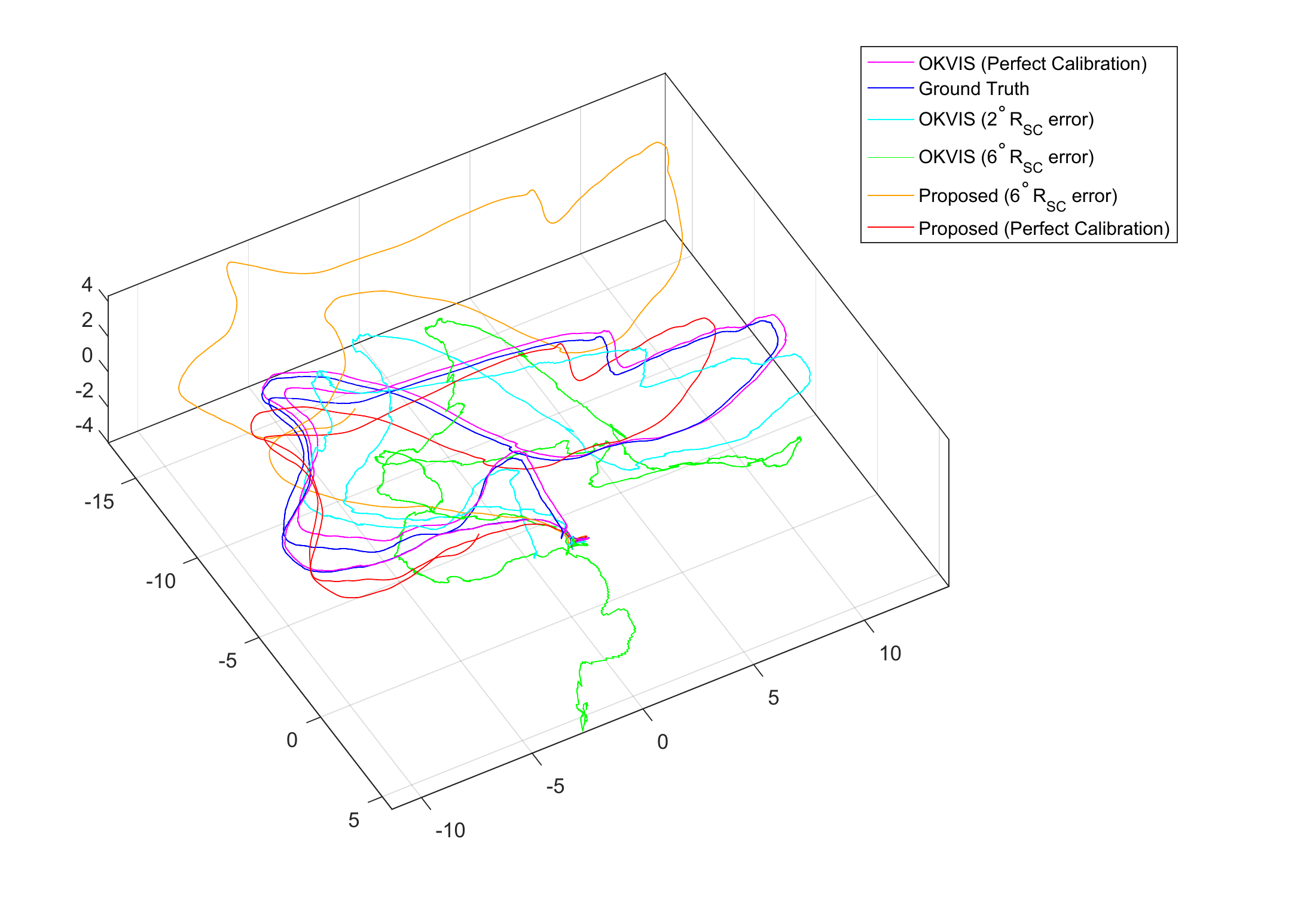}
\caption{6D MAV reconstructed trajectory using the proposed neural network compared to OK-VIS \cite{leutenegger2015keyframe}.}
\label{fig:euroc_trajectory}
\end{figure}
\begin{figure}[h!]
\centering
\includegraphics[width=\columnwidth]{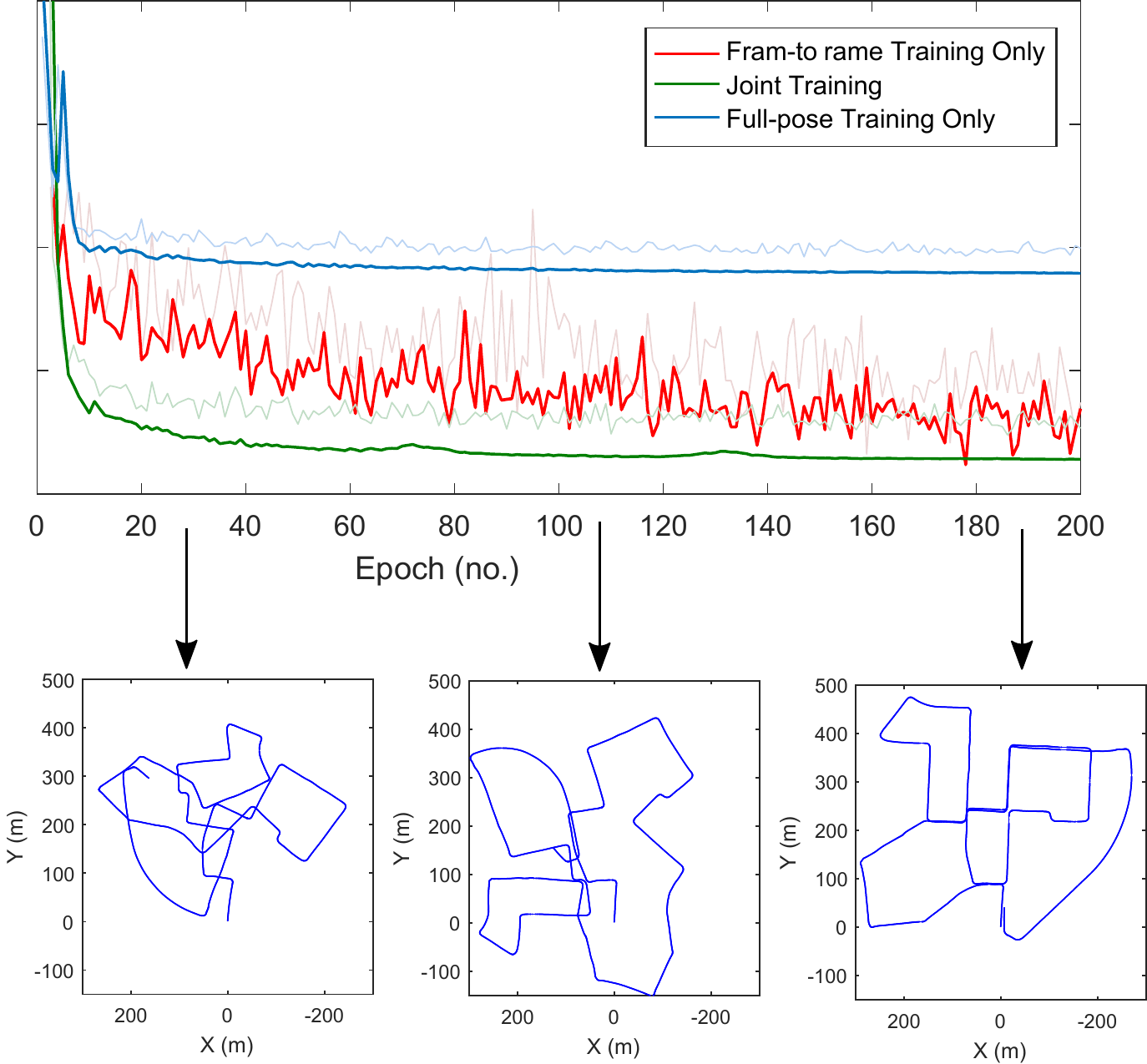}
\caption{Training performance of the network when training (1) only on the frame-to-frame transformations, (2) jointly on the $SE(3)$ layer and frame-to-frame transformation and (3) only on the concatenated pose. The training progress of the KITTI Seq-00 is shown for the best joint training.}
\label{fig:training_objective}
\end{figure}

 Fig. \ref{fig:euroc_trajectory} shows the comparison of the estimated MAV trajectory by OK-VIS and VINet for various levels of mis-calibration. It is evident that even when trained using no augmentation, the neural network degrades more gracefully in the face of mis-calibrated sensor data. 
\begin{table}[h]
\centering
\caption{Robustness of the VINet to sensor-camera calibration errors.}
\label{tbl:calibration}
\begin{tabular}{l|l|l|l|l}
                & $0^\circ$ & $5^\circ$ & $10^\circ$ & $15^\circ$ \\ \hline \hline
VI Net (no-aug) & 0.1751    & 0.8023    & 1.94       & 3.0671     \\ \hline
VI Net (w/ aug) & 0.1842    & 0.1951    & 0.2218     & 0.5178     \\ \hline
OK-VIS           & 0.1644    & 0.7916    & 1.9138     & FAILS         
\end{tabular}
\end{table}
Numerical results for the robustness test is shown in Table \ref{tbl:calibration}. VINet trained using no augmentation performs competitively compared to OK-VIS and does not fail with high calibration errors. The results for VINet trained using calibration augmentation show a significantly hindered decrease in accuracy as the calibration errors increase. This indicates that simply by training the network using mis-calibrated data, it can be made robust to mis-calibration errors. This property is unique to the network-based approach and rather surprising as it is very difficult, if not impossible, to increase the robustness of traditional approaches in this manner.
Time synchronization is another important calibration aspect which severely affects the performance of traditional methods. We tested VINet in this regard and found that it copes with time-synchronization error even better than extrinsic calibration error. When the streams are entirely unsynchronized, the IMU data are ignored and the network resorts to vision-only motion estimation.
The training performance in Fig. \ref{fig:training_objective} shows the difference between training solely on the F-2-F displacements, solely on the full $SE(3)$ pose and using our joint training method. The results show that joint training allows the network to converge more quickly towards low-error estimates over the training and validation sequences, while the F-2-F training converges very slowly and training on the full pose converges to a high-error estimate. 

\subsection{Autonomous driving: Structured Outdoor Trajectory} 
We further test the performance of VINet using the KITTI odometry benchmark \cite{geiger2012we,geiger2013vision}. The KITTI dataset comprises of 11 sequences collected from atop a passenger vehicle driving around a residential area with accurate ground-truth obtained from a Velodyne laser scanner and GPS unit. We use sequence 1-10 for training and 11 for testing. The monocular images and ground-truth are sampled at 10 Hz, while the IMU data is recorded at 100 Hz. Being recorded outdoors on structured roads, this dataset exhibits negligible motion blur and the trajectories follow very regular paths. However, it has different characteristics compared to the indoor EuRoC dataset which still make it very challenging. For example, the vehicle path contains many sharp turns and cluttered foliage areas which make data association between frames difficult for traditional visual odometry approaches. As the KITTI dataset does not provide tight-synchronization between the camera and IMU, we were unable to successfully run OK-VIS \cite{leutenegger2015keyframe} on this dataset. For comparison, we instead mimic the system of \cite{weiss2012real} by fusing pose estimates from LIBVISO2 \cite{geiger2013vision} with the inertial data using an EKF. We follow the standard KITTI evaluation metrics where we calculate the error for $100m, 200m, \dots, 500m$ sequences.

\begin{figure}[h!]
\centering
\includegraphics[width=1\columnwidth]{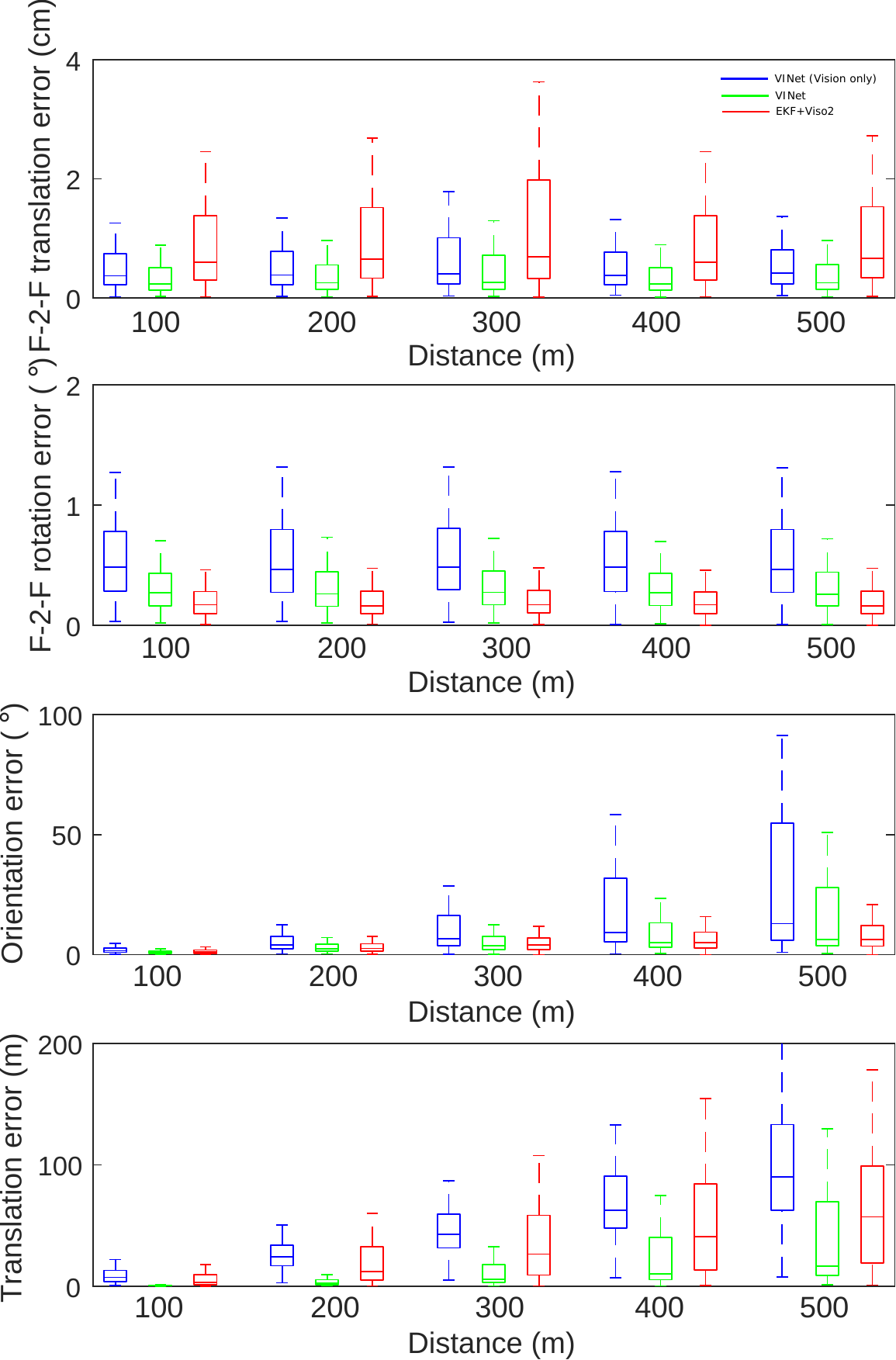}
\caption{Translation and orientation errors on the KITTI dataset. Method A: VNet (only image data), Method B: VINet (visual-inertial data), Method C: Viso2}
\label{fig:kitti_error}
\end{figure}

Fig. \ref{fig:kitti_error} shows the translation and orientation errors obtained on the KITTI dataset. As expected, the visual-inertial network (VINet) outperforms the network using visual data alone. VINet also outperms the VISO2 method in terms of translational error, however it suffers somewhat from estimating orientation where the IMU-Viso2 approach performs better. The high translational accuracy of VINet compared to its orientation estimation can possibly be attributed to its ability learn to predict scale from both the image data as-well as the IMU data which is not possible in traditional approaches. 

\section{Conclusion and Future Work}
In this paper we have presented VINet, an end-to-end trainable system for performing monocular visual-inertial aided navigation. We have shown that VINet performs on-par with traditional approaches which require much hand-tuning during setup. Compared to traditional methods, VINet has the key advantage of being able to learn to become robust to calibration errors. We believe that the VINet approach is a first step towards truly robust visual-inertial sensor fusion. For future work we intend to investigate the integration of VINet into a larger system with loop-closures and map-building as well as to perform a more in-depth analysis of the monocular VO and its ability to deal with the scale problem in abscene of the inertial data. 

\balance

\bibliography{refs}
\bibliographystyle{aaai}

\end{document}